\documentclass[journal]{IEEEtran}

\usepackage[latin1, utf8]{inputenc}
\usepackage{tikz}
\usepackage{fancyhdr}

\IEEEoverridecommandlockouts
\usepackage{pdfpages}
\usepackage{todonotes}
\usepackage{cite}
\usepackage{subfig}
\usepackage{svg}
\usepackage{placeins}
\usepackage{amsmath,amssymb,amsfonts}
\usepackage[ruled,linesnumbered]{algorithm2e}
\usepackage{graphicx}
\usepackage{textcomp}
\usepackage{multirow}
\usepackage{tablefootnote}
\usepackage{threeparttable}
\usepackage{blindtext}
\usepackage[inline]{enumitem}
\usepackage{xcolor}
\usepackage{layouts}

\bstctlcite{IEEEexample:BSTcontrol}
\usepackage{xcolor, bm}

\usepackage{gensymb}

\begin{document}

\author{
\IEEEauthorblockN{Soyed Tuhin Ahmed, Mehdi Tahoori}\\ \IEEEauthorblockN{Email: {soyed.ahmed, mehdi.tahoori}@kit.edu}\\
\IEEEauthorblockN{ITEC - Department of Computer Science, Karlsruhe Institute of Technology, Karlsruhe, Germany}}

\bstctlcite{IEEEexample:BSTcontrol}
\title{\huge Few-Shot Testing: Estimating Uncertainty of Memristive Deep Neural Networks Using One Bayesian Test Vector
}

\maketitle

\IEEEtitleabstractindextext{%
\abstract{
The performance of deep learning algorithms such as neural networks (NNs) has increased tremendously recently, and they can achieve state-of-the-art performance in many domains. However, due to memory and computation resource constraints, implementing NNs on edge devices is a challenging task. Therefore, hardware accelerators such as computation-in-memory (CIM) with memristive devices have been developed to accelerate the most common operations, i.e., matrix-vector multiplication. However, due to inherent device properties, external environmental factors such as temperature, and an immature fabrication process, memristors suffer from various non-idealities, including defects and variations occurring during manufacturing and runtime. Consequently, there is a lack of complete confidence in the predictions made by the model. To improve confidence in NN predictions made by hardware accelerators in the presence of device non-idealities, in this paper, we propose a Bayesian test vector generation framework that can estimate the model uncertainty of NNs implemented on memristor-based CIM hardware. Compared to the conventional point estimate test vector generation method, our method is more generalizable across different model dimensions and requires storing only one test Bayesian vector in the hardware. Our method is evaluated on different model dimensions, tasks, fault rates, and variation noise to show that it can consistently achieve $100\%$ coverage with only $0.024$ MB of memory overhead.
}

}

\section{Introduction}

Deep learning algorithms, particularly neural networks (NNs), have achieved remarkable performance improvements in recent years, becoming state-of-the-art in various domains. However, the implementation of NNs on edge devices poses significant challenges due to memory and computation resource constraints. Conventionally, NNs are implemented on von Neumann architectures, where the processing unit and memory are physically separated. Since NNs can have millions of parameters and inputs, communication between the processing unit and memory becomes a critical performance bottleneck, leading to the memory wall problem~\cite{keckler2011gpus}.

Hardware accelerator architectures have been developed to enable efficient execution of the most common operations in NNs, matrix-vector multiplication (MVM), by exploiting the inherent parallelism in the computation
\cite{chen2020survey}. However, computation-in-memory (CIM) architectures with emerging resistive non-volatile memories (NVMs) also offer the ability to implement the MVM operation directly within the memory where the data already reside~\cite{yu2018neuro}. Therefore, CIM can significantly reduce power consumption by overcoming the memory wall problem, making it suitable for edge devices. Furthermore, NVM technologies such as resistive random access memory (ReRAM)~\cite{wong2012metal}, phase change memories (PCMs)~\cite{burr2015experimental}, and spin-transfer-torque magnetic random access memory (STT-MRAM)~\cite{vincent2015spin} offer advantages that include zero leakage power, high endurance, and low read latency.

Despite their advantages, NVMs suffer from various non-idealities such as defects and variations arising during the manufacturing process and run-time operation, due to inherent device properties, external environmental factors such as temperature, and immature fabrication processes~\cite{Hofmann2014asyncflip2, chen2014rram, Park2010asyncflipRRAM,7116247, sun2019understanding,nair2020defect, niu2010impact,zhou2020noisy,kim2020efficient}. These non-idealities can negatively impact inference accuracy, reduce reliability, and introduce uncertainty in the predictions made by the NN models mapped into the hardware accelerators. Uncertain predictions are unacceptable for many applications, including safety-critical applications, where the consequences of an incorrect prediction can be catastrophic. Therefore, estimation followed by reduction of uncertainty is crucial for ensuring highly reliable operation. 

Existing approaches~\cite{Arjun2023BO, chen2021line, li2019rramedy, luo2019functional, soyed2023single} to estimate \emph{memristive-model uncertainty} typically rely on either \begin{enumerate*}[label={\alph*)},font={\color{black!50!black}\bfseries}]
    \item a large number of point estimates test vectors, 
    \item access to training data, or
    \item fine-tuning the model,
\end{enumerate*}
which can lead to high overhead in terms of computational and memory resources. 
Some studies implement Bayesian NNs (BNNs) in CIM~\cite{soyed_nanoarch22, ahmed_spindrop_2022, malhotra_exploiting_2020, soyed23DATE, yang2020all}. Although BNN can inherently estimate uncertainty, those approaches cannot estimate the uncertainty of conventional NNs and sometimes require changes to the common CIM architecture.

In this paper, we propose a Bayesian test vector generation framework that estimates the model uncertainty of conventional (point estimate) NNs implemented on a memristor-based CIM hardware accelerator (MHA). Our method does not require any changes to common CIM architectures, is generalizable across different model dimensions, does not require any access to training data, makes changes to the parameters of a pre-trained model, requires minimum test queries, and stores only one test vector in the hardware. We have extensively evaluated our method on different model dimensions, tasks, and non-ideal scenarios to show that our me can consistently achieve $100\%$ uncertainty estimation coverage. 


The remainder of the paper is organized as follows. In Section \ref{sec:background}, we provide some background for this work. Section \ref{sec:approach} details the proposed approach. Section \ref{sec:result} evaluates the proposed approach and discusses the results, followed by Section \ref{sec:conclusion}, which concludes the paper.

\section{Background}\label{sec:background}

\subsection{Memristive Devices and Sources of Uncertainty}\label{sec:memristor}

Emerging non-volatile memory (NVM) technologies, such as ReRAM, PCM, and STT-MRAM
exhibit two or more stable states depending on technology, which can be exploited for efficient storage and computation. For example, STT-MRAM has two stable resistance states, low resistance state (LRS) and high resistance state (HRS). However, ReRAM and PCM have multiple intermediate stable states. Nevertheless, memristive devices suffer from various types of faults and variations. These faults and variations can cause errors in the computed weighted sums and ultimately uncertainty in the NN predictions.


\paragraph{Faults}
Memristor devices can suffer from various faults. For example, faults caused by defects introduced during fabrication~\cite{nair2020defect, 7116247} and stress-induced failures~\cite{sun2019understanding} can cause some memristor call to be stuck in one of the stable states, e.g., a high- or low-resistance state. On the other hand, retention faults~\cite{Hofmann2014asyncflip2, Park2010asyncflipRRAM}, write faults~\cite{7116247}, and read-disturb faults~\cite{chen2014rram} can randomly flip the state of a memristor. That means that a cell in the HRS state can randomly flip to the LRS state and vice versa. The same phenomenon is likely to occur in a multilevel cell. Also, the effect of a stuck-at-fault is observed only when a cell is stuck in a state different from its original state, similar to a state flip. Therefore, we have modelled these faults into two types: bit-flip and level-flip faults. In the bit flip fault model, individual bits of weights are randomly flipped. However, in the level-flip fault model, the state of each weight is randomly flipped.

\begin{figure*}
    \centering
    \includegraphics[width=\linewidth]{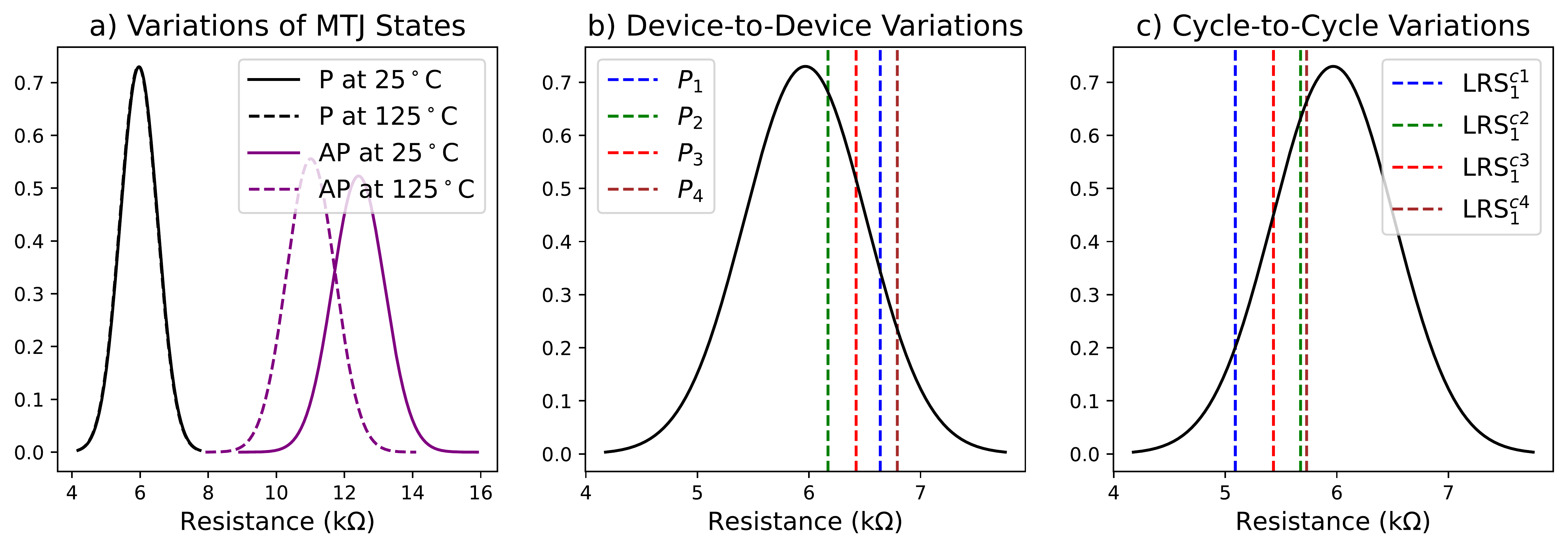}
    \caption{a) Variations of MTJ devices due to manufacturing and thermal variations, b) variations of four MTJs ($P_1\cdots P_4$) in the P-state. The respective variations of the devices are randomly sampled from their distribution, and c) variations in resistance of a single RRAM device in LRS state with different programming cycles $c_1\cdots c_4$.}
    \label{fig:d2d}
\end{figure*}

\paragraph{Variations}
Due to the manufacturing variation (spatial variation), the conductance of a memristor is usually represented as a distribution rather than as a single point~\cite{niu2010impact}. The parameters of the distribution, e.g., mean and variance for a Gaussian distribution, vary depending on NVM technology. In STT-MRAM, the distribution is represented as a Gaussian distribution, as shown in Fig.~\ref{fig:d2d}.
Nevertheless, due to manufacturing variation, a memristor device can exhibit different resistance even though it is in the same state. This phenomenon is known as device-to-device variations and is conceptually depicted in Fig.~\ref{fig:d2d} (a). On the other hand, in certain NVM technologies, e.g., RRAM, the resistance of a device can vary during each programming cycle, also known as cycle-to-cycle variations. The concept of cycle-to-cycle variation is depicted in Fig.~\ref{fig:d2d} (b).

According to the variation modelling proposed in \cite{kim2020efficient}, variations in NVM technologies, in general, can be categorized as additive and multiplicative variations. Additive variation is independent of the magnitude of conductance, but multiplicative variation depends on it. Due to external factors such as temperature, the conductance distribution can also fluctuate with time (temporal variation)~\cite{zhou2020noisy}. We have modelled both types of variation as Gaussian distributions. We have also considered both cycle-to-cycle and device-to-device variations in the fault model as we sample independently for each weight. 



\subsection{Neural Networks}

(Artificial) neural networks (NNs) are computational models inspired by the biological neurons in the human brain. NNs can recognize patterns from training data and make decisions given unseen inference data. There are various types of NN, but convolutional neural networks (CNNs) are the most efficient for image, audio, and video data.

The basic computation of NNs involves a series of layers, each performing a weighted sum of its inputs, followed by a non-linear activation function applied (element-wise) to the weighted sum. 
The overall output of a NN ($\mathcal{F}$) can be summarized as follows:
\begin{equation}
    y = \mathcal{F}_\theta(\mathbf{x}).
\end{equation}
Here, $\theta$ represents all the learnable parameters of NN such as weights and biases and $\mathbf{x}$ is the input vector.

To reduce memory requirements and computation requirements, usually $\theta$ is quantized to lower bit precision. Due to the limited stable states of NVMs, quantization of  $\theta$  is beneficial. Otherwise, inference accuracy may degrade due to quantization error. In this work, we quantize the weights of NNs into 8 bits using post-training quantization. 


\subsection{Computation-in-Memory}
\begin{figure}
    \centering
    \includegraphics[,scale=0.55]{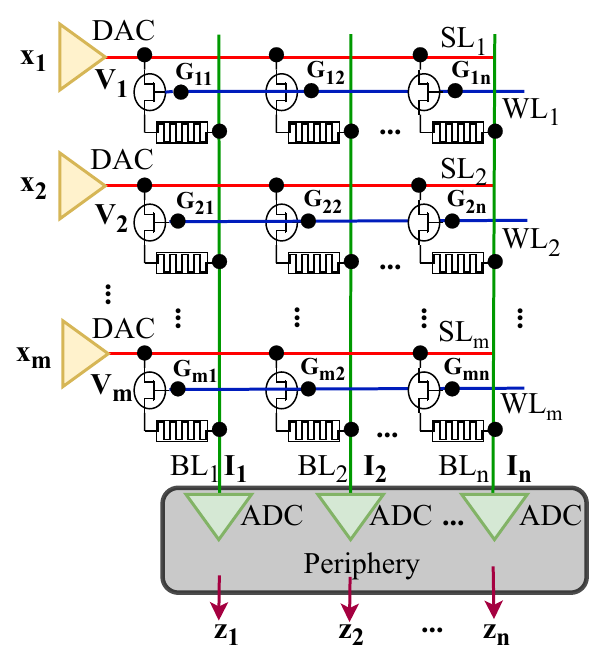}
    \caption{Crossbar array of CIM architecture. Each crosspoint consists of a memristor with a 1TIM configuration.}
    \label{fig:crossbars}
\end{figure}

In CIM architectures, memristive devices are arranged in two-dimensional crossbar arrays with rows representing input neurons and columns representing output neurons (see Fig.~\ref{fig:crossbars}). Each memristor cell is arranged in a 1-transistor, 1-memory (1TIM) configuration. 1T1R configuration facilitates selecting the memristor to be programmed to the desired conductance state, and the transistor aids in limiting the sneak-path currents in the crossbar. Usually, the 8-bit quantized weights of the pre-trained NN are mapped to the crossbar arrays by programming the individual memristive devices into high and low resistance states~\cite{chen2021line}. Specifically, low-resistance states encode "$0$" and high-resistance states encode "$1$".

A digital-to-analog converter (DAC) converts input neuron values into voltages and apply to the rows of the crossbar array. Consequently, currents flow through each memristor, as stated by Ohm's law. The current sum $\mathbf{I}_{out}$ in each column, $\mathbf{I}_{out} = \mathbf{V}_{in} \mathbf{G}$,
represents the weighted sum operation of a layer. where $\mathbf{V}_{in}$ is the input voltage vector, and $\mathbf{G}$ represents the conductance matrix of the memristive devices in the crossbar array. Sensing circuitry, such as an analog-to-digital converter (ADC), is employed to convert the weighted sums in the crossbar array into digital values. 

\subsection{Uncertainty In Deep Learning}

Uncertainty 
in deep learning
represents the degree of confidence or lack thereof in the predictions made by trained NNs. There are two types of uncertainties: aleatoric and epistemic. Aleatoric uncertainty, also called data uncertainty, corresponds to the intrinsic variability and randomness in the data. Aleatoric uncertainties are irreducible. On the other hand, epistemic uncertainty, also called model uncertainty, refers to the ambiguity in the predictions made by a trained neural network due to the limitations in the model's architecture, training data, or other factors that affect its generalization capabilities. Model uncertainties are reducible with more information.

In the context of memristive NNs, a special kind of model uncertainty is introduced. We refer to it as \emph{memristive model uncertainty}. The sources of uncertainty were discussed earlier in Section~\ref{sec:memristor}. These non-idealities can impact the accuracy and reliability of the NN computations, leading to uncertain predictions. 

As with every other aspect of life, uncertainty is avoided whenever possible. The presence of memristive model uncertainty can be challenging for safety-critical applications, where the reliability of the predictions is of utmost importance. It is crucial to accurately estimate and account for memristive model uncertainty when designing and deploying  NNs to MHA. Developing methods to estimate memristive model uncertainty in the presence of memristive non-idealities can help improve the robustness and trustworthiness of these accelerators in real-world applications.



\subsection{Related Works}
Bayesian NNs (BNNs) are a type of stochastic NNs that can inherently estimate uncertainty in each prediction. Several works~\cite{soyed_nanoarch22, ahmed_spindrop_2022, malhotra_exploiting_2020, soyed23DATE, yang2020all} have been proposed to implement BNNs to CIM hardware for estimating uncertainty. However, implementing BNNs to CIM requires changes to CIM hardware. On the contrary, our method can estimate the uncertainty of conventional NNs without requiring any changes to CIM architectures.

Recent work A. Chaudhuri et al.~\cite{Arjun2023BO} proposed to generate functional test patterns using a black-box optimization method for the systolic array-based accelerator. They use Bayesian optimization to generate targeted test images for stuck-at faults. Their method can achieve high fault coverage, but their method is evaluated on the primitive NN model LeNet-5, an easier MNIST dataset, and single stuck-at faults. Their method may not be scalable for larger NN models, harder datasets, and memristor-specific non-idealities such as variations. 

Several approaches with point estimate test vectors such as \cite{chen2021line, li2019rramedy, luo2019functional} have been proposed. These methods involve analyzing the deviation in inference accuracy when using original training data or synthetic testing data in the presence of faults. Li et al. \cite{li2019rramedy} proposed a method to generate synthetic testing data using the fast gradient sign method (FGSM), an adversarial example generation method. Their method, called "pause-and-test", may result in prolonged system downtime as a result of utilizing numerous test vectors. Chen et al.~\cite{chen2021line} proposed watermarking of training data, followed by retraining of the neural network (NN) on those data to introduce a backdoor. While these methods effectively identify deviations, they are invasive as they require changes to NN parameters, which many users may not allow as they view the pre-trained NN as intellectual property. Furthermore, this method requires a significant amount of testing data and on-chip storage (depending on the availability of on-chip retraining data), as shown later in Section~\ref{sec:comp_rel}.


Some hardware-based approaches have also been proposed. Liu et al.~\cite{9693118} proposed implementing adder trees to continuously monitor the dynamic power consumption of memristive crossbar arrays. However, implementing adder trees introduces additional hardware overhead.

In contrast, our Bayesian test vector generation framework and respective uncertainty estimation approach
\begin{enumerate*}[label={\alph*)},font={\color{black!50!black}\bfseries}]
\item are not only non-invasive but also do not require access to in-distribution data such as training or validation data,
\item are generalizable across different NN models,
\item needs negligible overhead in terms of storage and power, 
\item needs minimum forward passes, which leads to minimum latency, 
\item can consistently achieve high coverage on different fault rates or variation scales.
\end{enumerate*}



\section{Uncertainty Estimation of Memristive NNs}\label{sec:approach}
\subsection{Problem Statement and Motivation}
\begin{figure}
\centering
\includegraphics[width=\linewidth]{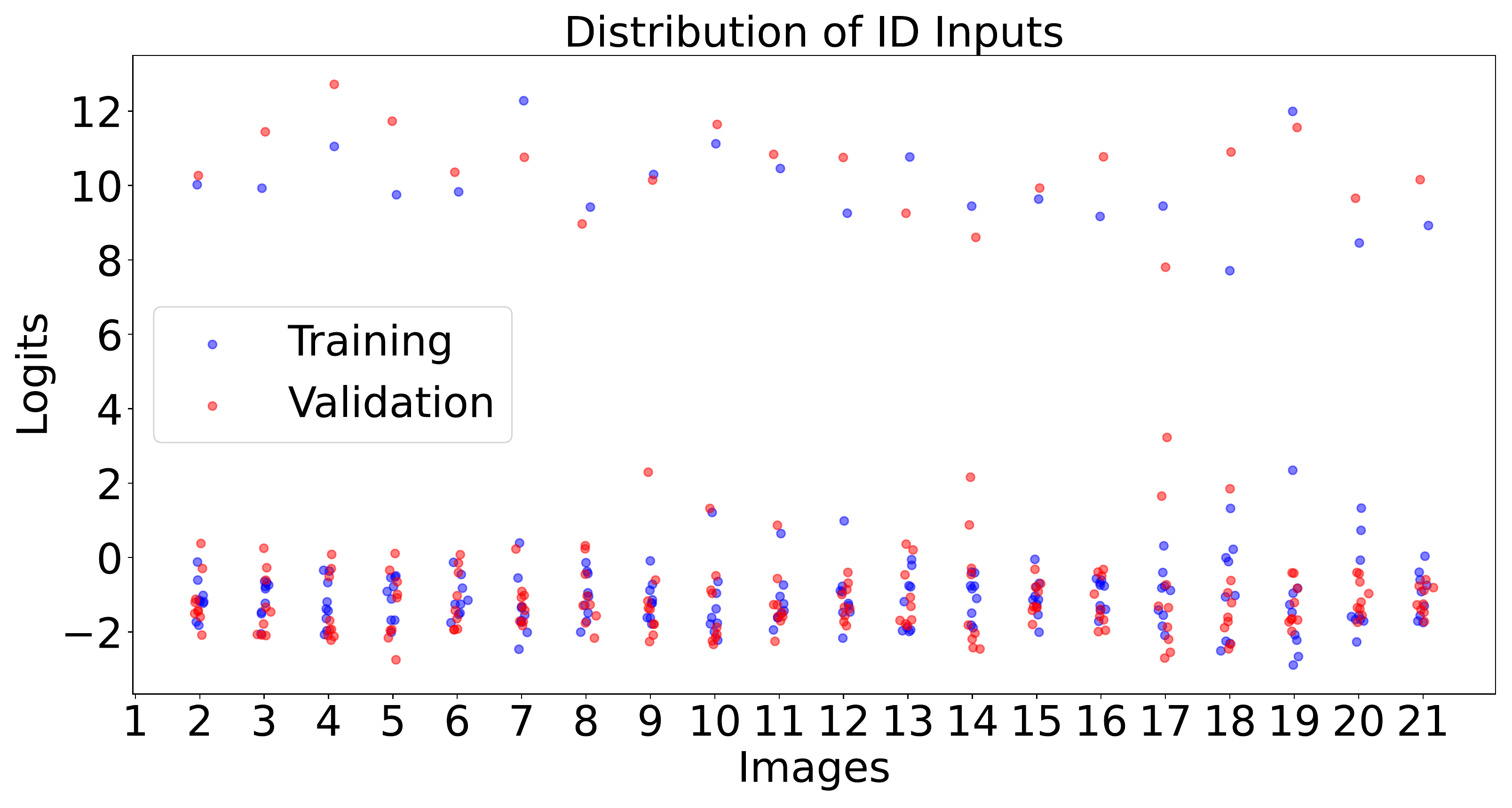}
\caption{Distribution of logits on the fault- and variation-free RepVGG~\cite{ding2021repvgg} model on the CIFAR-10 dataset when several randomly sampled inputs from training and validation are applied. The distribution logits change from one input to another. }
\label{fig:obeservation_baseline}
\end{figure}

An NN with parameters $\theta$ gives the output $y$ based on the input. Parameters $\theta$ are learned so that the predicted output $y$ is close to ground truth $\hat{y}$. Therefore, even though one has access to in-distribution data, e.g., training or validation data, their respective output \begin{enumerate*}
    \item would change from one input to the next,
    \item one model to the next, and
    \item uncertainty may not be minimum, see Fig.~\ref{fig:obeservation_baseline}.
\end{enumerate*}

\begin{figure}
\centering
\includegraphics[width=\linewidth]{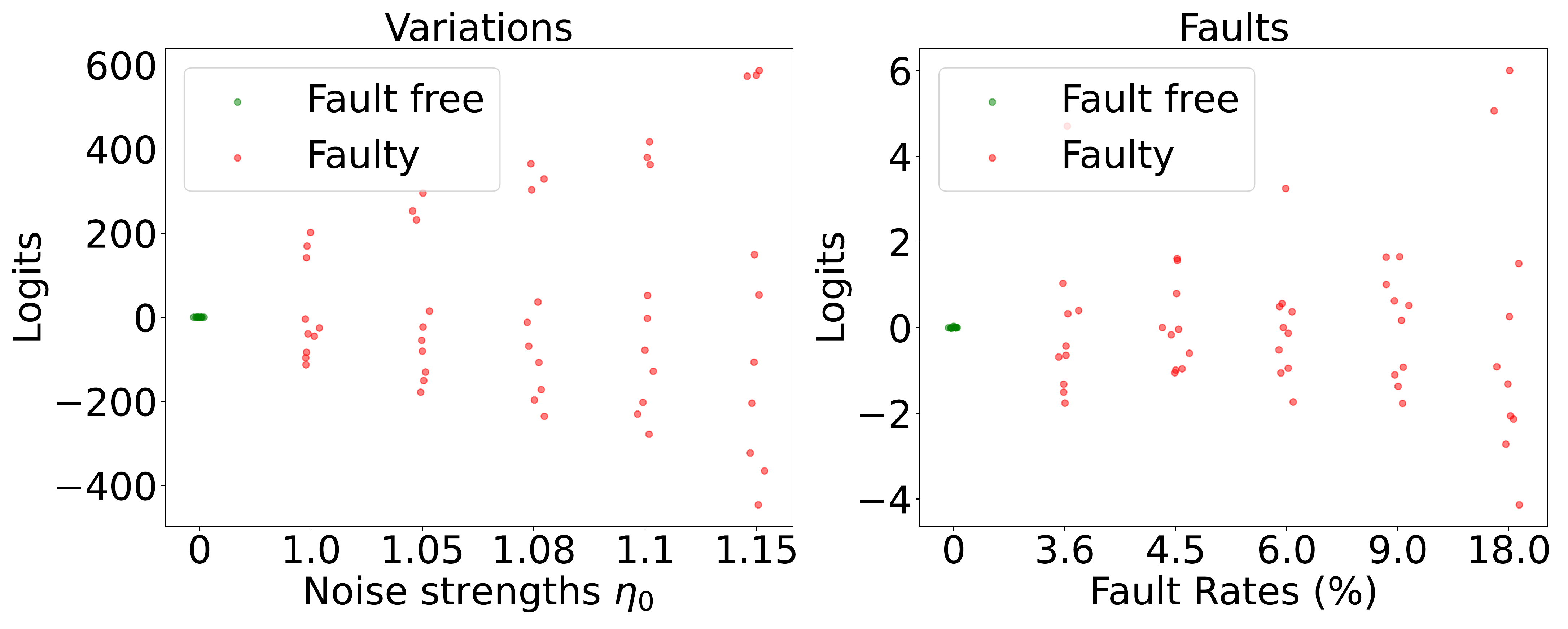}
\caption{Change in output distribution of logits of RepVGG~\cite{ding2021repvgg} model on the CIFAR-10 dataset due to a) variations and b) faults. The spread among logits increases as the noise scale of variations and fault rate increases.  }
\label{fig:baseline_changes}
\end{figure}

On the other hand, we have observed that the spread between logits of a NN (in general) increases as faults and variations in its parameter increase, as depicted in Fig.~\ref{fig:baseline_changes}. Based on this observation, we propose to measure the standard deviation $\sigma_y$ of the logits of a \emph{model under test} (MUT) to estimate the \emph{memristive model uncertainty}.

\begin{figure*}
\centering
\includegraphics[width=\linewidth]{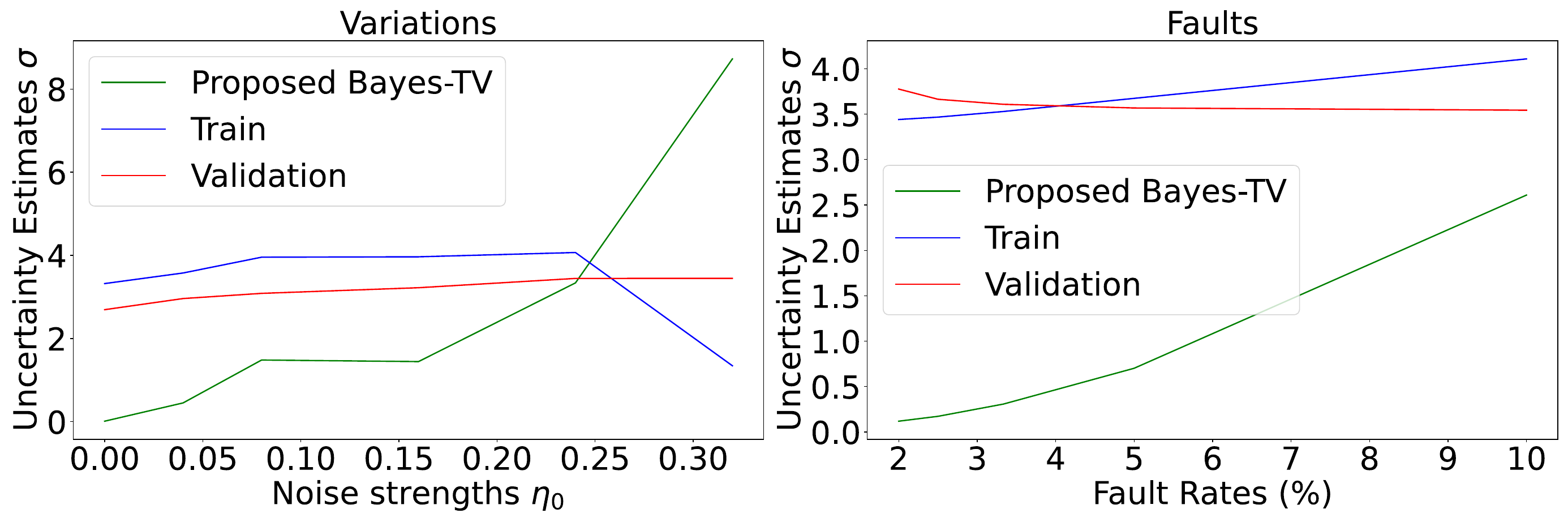}
\caption{Relative sensitivity of uncertainty estimates given proposed Bayesian test vector input as well as randomly sampled training and validation. The change in uncertainty estimates is much higher for our proposed Bayesian test vector.  }
\label{fig:senstitivity_prop}
\end{figure*}

\begin{figure*}
\centering
\includegraphics[width=\linewidth]{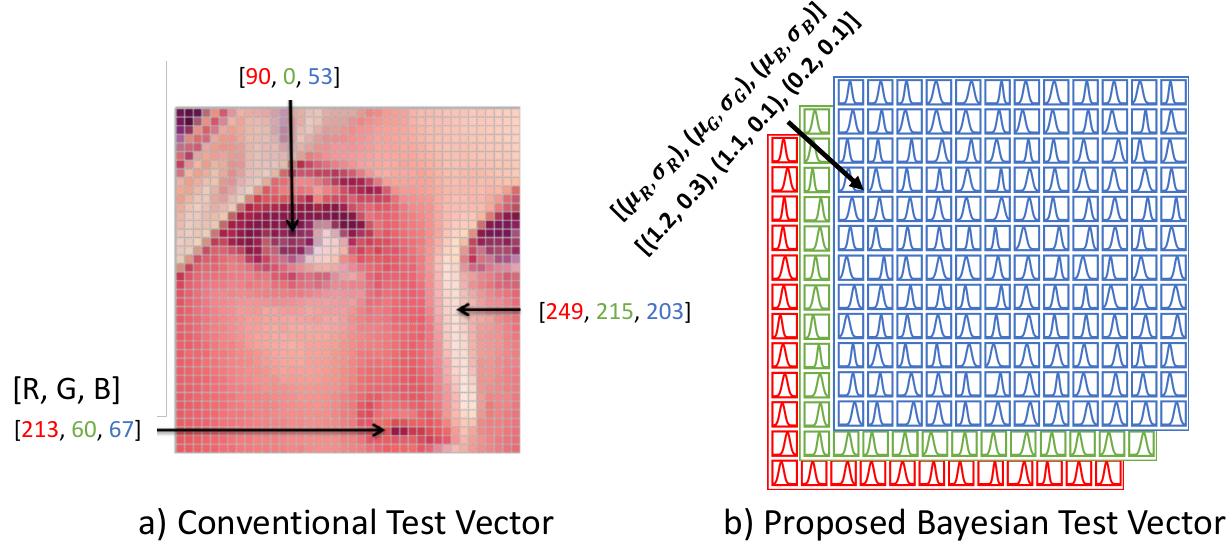}
\caption{An example of a) conventional test vector with each pixel representing \emph{single point value} for the Red, Green, and Blue (RGB) channels, and b) proposed Bayesian test vector with each pixel representing \emph{an independent distribution} for the RGB channels. }
\label{fig:test_vect_visual}
\end{figure*}

\begin{figure}
    \centering
    \includegraphics[width=0.9\linewidth]{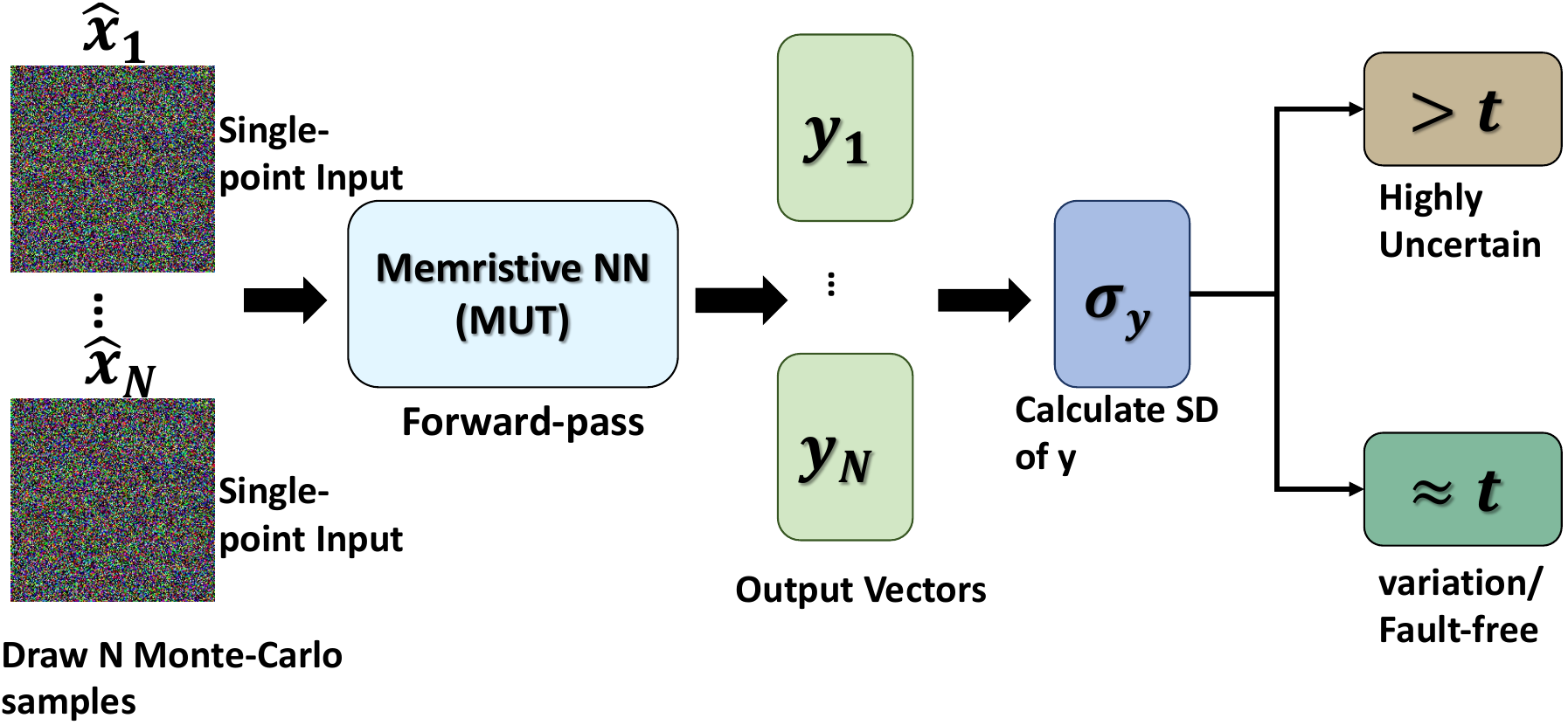}
    \caption{Flowchart of the proposed uncertainty estimation method for a model under test (MUT). If the standard deviation (SD) of the output of MUT $\sigma_y$ is higher than a pre-defined threshold $t$, then MUT is highly uncertain. }
    \label{fig:flow_prop}
\end{figure}

To achieve the desired $\sigma_y$, we proposed a single Bayesian test vector generation framework. In a Bayesian test vector, each element of the input, e.g., pixels for classification tasks, is a distribution rather than a single-point estimate value, as shown in Fig.~\ref{fig:test_vect_visual}. As a result, one test vector can be learned and $N = {1\cdots N}$ Monte Carlo samples (MC samples) can be taken from it. In each MC sampling step, element-wise sampling is performed, resulting in an input vector with a single value element. Each of the MC samples is forward passed $N$ times through the MUT to create an output distribution for the calculation of uncertainty estimates $\sigma_y$ (see Fig.~\ref{fig:flow_prop}). Here, $N$ depends on the number of neurons $C$ in the final layers. A MUT with a large $C$, for example, $C\ge100$, has enough elements in the output vector to calculate the uncertainty with one sample and forward pass. However, a model under test with a small $C$, for example, a regression or binary classification task with $C=1$, will require multiple MC samples from the distribution of the Bayesian test vector and forward passes to calculate the memristive model uncertainty.

Our proposed Bayesian test vector can give a distinguished output in the case of memristive model uncertainty. In the case of a memristive NN that is fault-free and variation-free, in an ideal scenario, the uncertainty $\sigma_y$ is minimum. Consequently, the uncertainty measures $\sigma_y$ of a model given our proposed Bayesian test vector is much more sensitive compared to training and validation data, as demonstrated by Fig.~\ref{fig:senstitivity_prop}. Therefore, our Bayesian test vector can potentially estimate uncertainty at low fault rates or variations.

Although for a certain MUT, several forward passes $N$ are required to estimate the uncertainty, the overhead is still minimal. This is because, in deep learning, one of the factors that determines the difficulty of a task is the number of classes in it. A difficult task, for example, ImageNet-1k classification~\cite{deng2009imagenet}, usually requires a larger NN model, while an easier task requires a smaller NN model. In our approach, we take several samples $N$ and forward-pass them through only on a small MUT. However, for a large MUT, we only take a single sample and forward-pass it through the MUT.

\subsection{Optimization Process of The Bayesian Test Vector}

To obtain the test vector $\hat{x}$ we employ Bayesian inference. It uses Bayes' theorem to update our belief about the parameter of interest (in this case, the test vector $\hat{x}$) given some observed data (in this case, the output $y$). It can be written as:

\begin{equation}
P(\hat{x}|y) = \frac{P(y|\hat{x})P(\hat{x})}{P(y)}.
\end{equation}
Where, $P(\hat{x}|y)$ is the posterior distribution of the test vector given the output, $P(y|\hat{x})$ is the likelihood of the output given the test vector, $P(\hat{x})$ is the prior distribution over the test vector, and $P(y)$ is the evidence or marginal likelihood of the output.

In our Bayesian optimization setting, the primary objective is to find the configuration of the test vector, $\hat{x}$, that maximizes the posterior distribution, $P(\hat{x}|y)$, given the output, $y$. However, since the direct computation of the posterior is often intractable, we resort to variational inference techniques to approximate our true posterior with a variational distribution, $q(\hat{x}|y)$, that comes from a simpler or more tractable family of distributions. Hence, we choose a Gaussian distribution as the variational distribution that is parameterized by mean $\mu_{\hat{x}}$ and standard deviation $\sigma_{\hat{x}}$.

Our objective can therefore be rephrased as finding the $\hat{x}$ that maximizes $q(\hat{x}|y)$, which can be considered as an approximation of the expected lower bound (ELBO). This is because, in the context of variational inference, the ELBO is the objective function that we aim to maximize. Hence, its negative can be seen as a loss function that needs to be minimized. The ELBO is defined as:

\begin{equation}
\text{ELBO} = \mathbb{E}_{q(\hat{x}|y)}[\log P(y|x)] - \text{KL}(q(\hat{x}|y) || P(\hat{x}))
\end{equation}

Where the first term, $\mathbb{E}_{q(\hat{x}|y)}[\log P(y|x)]$, is the expected log-likelihood of the data to observe the desired output $y$, and the second term, $\text{KL}(q(\hat{x}|y) || P(\hat{x}))$, is the Kullback-Leibler (KL) divergence between the variational distribution and the prior, which can be thought of as a regularization term that makes sure that the variational distribution is close to the prior.

We have rephrased the objective function for our need as follows:

\begin{equation}
\text{Loss} = - \text{ELBO} =  \sqrt{\frac{\sum{(y-\mu_y)^2}}{C}} + \alpha \times \text{KL}(q(\hat{x}|y) || P(\hat{x}))
\end{equation}

Here, the term $\sqrt{\frac{\sum{(y-\mu_y)^2}}{C}}$ serves as a proxy for the likelihood term. It is designed to encourage the optimization process to find an $\hat{x}$ that results in $y$ with as low a standard deviation as possible. This term essentially computes the root mean squared deviation of the outputs from their mean, reflecting our goal of minimizing output uncertainty. Furthermore, we introduce a hyperparameter $\alpha$ that can control the strength of the regularization term, $\text{KL}(q(\hat{x}|y) || P(\hat{x}))$.

The optimization process involves iterative adjustment of $\mu_{\hat{x}}$ and $\sigma_{\hat{x}}$ of $q(\hat{x}|y)$ elementwise to minimize the loss function. This can be achieved using a gradient-based optimization algorithm, such as stochastic gradient descent (SGD). At each step, we evaluate the loss and adjust the parameters of $q(\hat{x}|y)$ elementwise in the direction that reduces the loss.

The outcome of the proposed optimization process is a test vector $\hat{x}$ that, when used with a MUT, produces an output vector with minimized uncertainty $\sigma_y$, thereby achieving our goal of a distinguished output in the presence of memristive model uncertainty.



\subsection{Application of Proposed Uncertainty Estimation Method}
Our proposed method can estimate the uncertainty of memristive NNs at a given time during post-mapping, but before deployment of the NN and during online operation. If the model is uncertain, that is, the uncertainty estimates $\sigma_y$ is more than a pre-defined threshold $t$, the uncertainty reduction method can be used. Common uncertainty reduction methods are re-training~\cite{yoshikiyo2023nn} and re-calibration~\cite{ahmed2023fault}. After uncertainty reduction methods are applied, the uncertainty of the model is re-estimated for verification. If the memristive uncertainty is satisfactorily low, then the model resumes normal operation. Otherwise, a more sophisticated uncertainty reduction method or hardware replacement may be required before the model can resume normal operation. 

\begin{figure}
    \centering
    \includegraphics[width=0.9\linewidth]{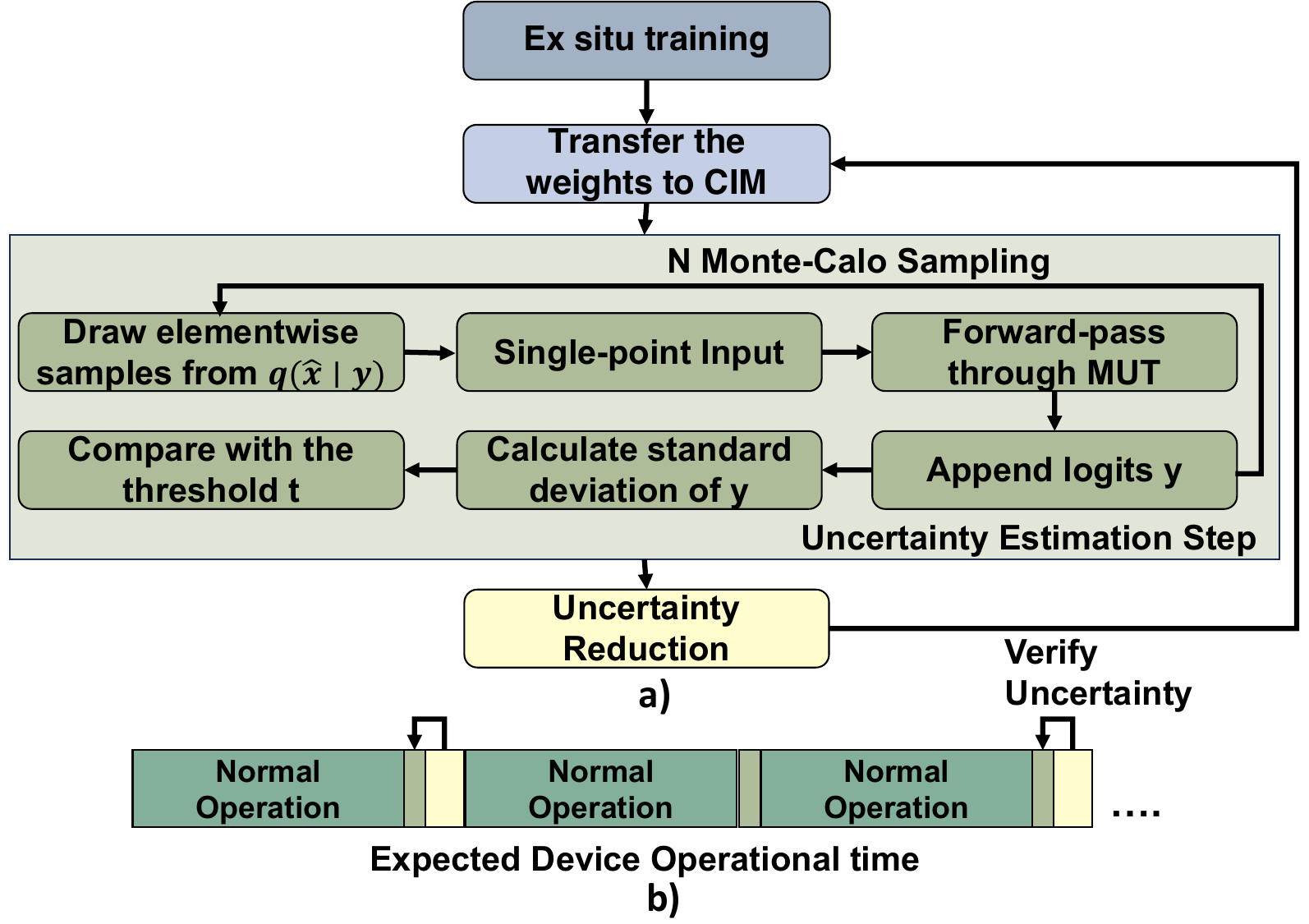}
    \caption{Flowchart of the application of the proposed uncertainty estimation method during a) post-mapping but pre-deployment, b) post-deployment (online) operation. }
    \label{fig:flow_application}
\end{figure}

The overall flow for the application of our memristive uncertainty estimation approach is depicted in Fig.~\ref{fig:flow_application}. Latency for uncertainty estimation during pre-deployment may not be important as NN is not in operation. However, it is highly important during online operation, as normal NN operation is paused while the uncertainty estimation process is carried out.
Our uncertainty estimation approach is designed to keep this latency to a minimum. Consequently, our method improves the confidence and reliability of the prediction.

\begin{table*}[]
\caption{Summary of the evaluated models.}
\resizebox{\linewidth}{!}{
\begin{tabular}{|cccccc|}
\hline
\multicolumn{1}{|c|}{Model}               & \multicolumn{1}{c|}{Accuracy}  & \multicolumn{1}{c|}{\# of Params.} & \multicolumn{1}{c|}{Layers} & \multicolumn{1}{c|}{Dataset}                      & Input Shape                     \\ \hline\hline
\multicolumn{6}{|c|}{Classification (Supervised Learning)}                                                                                                                                \\ \hline
\multicolumn{1}{|c|}{ResNet-20~\cite{he2016deep}}           & \multicolumn{1}{c|}{$92.60\%$} & \multicolumn{1}{c|}{$0.27\times10^6$} & \multicolumn{1}{c|}{20}     & \multicolumn{1}{c|}{\multirow{2}{*}{CIFAR-10~\cite{krizhevsky2009learning}}}    & \multirow{2}{*}{$32\times32$}   
\\
\multicolumn{1}{|c|}{RepVGG-A0~\cite{ding2021repvgg}}           & \multicolumn{1}{c|}{$94.39\%$} & \multicolumn{1}{c|}{$7.84\times10^6$} & \multicolumn{1}{c|}{22}     & \multicolumn{1}{c|}{}                             &                                 
\\ \hline
\multicolumn{1}{|c|}{ResNet-56~\cite{he2016deep}}           & \multicolumn{1}{c|}{$72.63\%$} & \multicolumn{1}{c|}{$0.86\times10^6$} & \multicolumn{1}{c|}{56}     & \multicolumn{1}{c|}{\multirow{2}{*}{CIFAR-100~\cite{krizhevsky2009learning}}}   & \multirow{2}{*}{$32\times32$}   \\
\multicolumn{1}{|c|}{MobileNet-V2~\cite{sandler2018mobilenetv2}}        & \multicolumn{1}{c|}{$74.20\%$} & \multicolumn{1}{c|}{$2.35\times10^6$} & \multicolumn{1}{c|}{53}     & \multicolumn{1}{c|}{}                             &                                 \\ \hline
\multicolumn{1}{|c|}{Inception-V3~\cite{szegedy2016rethinking}}        & \multicolumn{1}{c|}{$77.29\%$} & \multicolumn{1}{c|}{$27.2\times10^6$} & \multicolumn{1}{c|}{48}     & \multicolumn{1}{c|}{\multirow{2}{*}{ImageNet-1k~\cite{deng2009imagenet}}} & \multirow{2}{*}{$224\times224$} \\
\multicolumn{1}{|c|}{DenseNet-201~\cite{huang2017densely}}        & \multicolumn{1}{c|}{$76.89\%$} & \multicolumn{1}{c|}{$20.0\times10^6$} & \multicolumn{1}{c|}{201}    & \multicolumn{1}{c|}{}                             &                                 \\ \hline
\multicolumn{6}{|c|}{Semantic Segmentation (Supervised Learning)}                                                                                                                                                                      \\ \hline
\multicolumn{1}{|c|}{UNnet~\cite{ronneberger2015u}}               & \multicolumn{1}{c|}{$98.75\%$} & \multicolumn{1}{c|}{$7.76\times10^6$} & \multicolumn{1}{c|}{23}     & \multicolumn{1}{c|}{Brain-MRI~\cite{buda2019association}}                    & $224\times224$                  \\ \hline
\multicolumn{1}{|c|}{FCN~\cite{long2015fully}}     & \multicolumn{1}{c|}{$91.40\%$}  & \multicolumn{1}{c|}{$35.3\times10^6$} & \multicolumn{1}{c|}{57}     & \multicolumn{1}{c|}{MS COCO~\cite{lin2014microsoft}}                      & $224\times224$                  \\ \hline
\multicolumn{6}{|c|}{Generative Method (Unsupervised Learning)}                                                                                                                                                                        \\ \hline
\multicolumn{1}{|c|}{DCGAN-G~\cite{radford2015unsupervised}}     & \multicolumn{1}{c|}{-}         & \multicolumn{1}{c|}{$3.74\times10^6$} & \multicolumn{1}{c|}{5}      & \multicolumn{1}{c|}{\multirow{2}{*}{FASHIONGEN~\cite{rostamzadeh2018fashion}}}  & $1\times120$                    \\
\multicolumn{1}{|c|}{DCGAN-D~\cite{radford2015unsupervised}} & \multicolumn{1}{c|}{-}         & \multicolumn{1}{c|}{$2.93\times10^6$} & \multicolumn{1}{c|}{5}      & \multicolumn{1}{c|}{}                             & $64\times64$                    \\ \hline
\end{tabular}
}
\label{tab:models}
\end{table*}


\subsection{Sampling From Bayesian Test Vector}
We store element-wise $\mu_{\hat{x}}$ and $\sigma_{\hat{x}}$ of the Gaussian distribution (variational distribution) in hardware. Element-wise sampling during the training and uncertainty estimation step is done as follows:

\begin{equation}
    \text{{samples}} = \mathcal{N}(0, 1) \times \sqrt{\exp({\sigma_{\hat{x}}}}) + {\mu_{\hat{x}}}.
\end{equation}

where $\mathcal{N}(0,1)$ is a unit Gaussian distribution that can be implemented in software or hardware. The expression $\sqrt{\exp({\sigma_{\hat{x}}}})$ calculates the standard deviation from the logarithm of the variance. It ensures numerical stability and avoids numerical underflow issues. The proposed element-wise sampling is inspired by the re-parameterization trick proposed by Kingma and Welling~\cite{kingma2013auto}.



\section{Evaluation of the Proposed Approach}\label{sec:result}

\subsection{Simulation Setup}

\subsubsection{Evaluated Models} Our method is evaluated on models from different deep learning paradigms, specifically classification, semantic segmentation, and generative methods with different state-of-the-art (SOTA) models~\cite{he2016deep, ding2021repvgg, sandler2018mobilenetv2, szegedy2016rethinking, huang2017densely, ronneberger2015u, long2015fully, radford2015unsupervised} with up to $201$ layers. The number of classes in the benchmark datasets~\cite{krizhevsky2009learning, deng2009imagenet} varies from $10$ to $1000$ for the classification tasks. The generator and discriminator models of the Convolutional Generative Adversarial Network (DCGAN)~\cite{radford2015unsupervised} are evaluated separately.

All pre-trained models were downloaded from the PyTorch Hub library and the GitHub repository~\cite{Yaofo_Chen}. Therefore, no changes to the training procedure are made. Our Bayesian test vector generation process did not alter the model parameters. Thus, the black-box nature of our method remains consistent. For the hyperparameter $\alpha$, a value between $10^3-10^8$ is chosen.

Furthermore, we evaluated our approach on a range of input shapes, from $1\times120$ to $224\times224$. Table~\ref{tab:models} summarized all models, their baseline accuracy, the number of parameters, layers, and the shape of the input image. All models in Table~\ref{tab:models} are evaluated with a single MC sample and forward pass, $N=1$. Therefore, the uncertainty estimation is done in a \textbf{single shot}.

\begin{table*}
\caption{Uncertainty estimation coverage for different NN models and datasets under varying noise strengths for both multiplicative and additive variations.}\label{tab:var}
\resizebox{\linewidth}{!}{
\begin{tabular}{|cccccccccccc|}
\hline
\multicolumn{1}{|c|}{\multirow{2}{*}{Model}} & \multicolumn{1}{c|}{\multirow{2}{*}{Dataset}}   & \multicolumn{5}{c|}{Multiplicative Variations}                                                                                                 & \multicolumn{5}{c|}{Additive Variations}                                                                                  \\ \cline{3-12} 
\multicolumn{1}{|c|}{}                       & \multicolumn{1}{c|}{}                           & \multicolumn{1}{c|}{$\eta_0^1$}  & \multicolumn{1}{c|}{$\eta_0^2$}  & \multicolumn{1}{c|}{$\eta_0^3$}  & \multicolumn{1}{c|}{$\eta_0^4$}  & \multicolumn{1}{c|}{$\eta_0^5$}  & \multicolumn{1}{c|}{$\eta_0^1$}  & \multicolumn{1}{c|}{$\eta_0^2$}  & \multicolumn{1}{c|}{$\eta_0^3$}  & \multicolumn{1}{c|}{$\eta_0^4$}  & $\eta_0^5$  \\ \hline\hline
\multicolumn{12}{|c|}{Classification (Supervised Learning)}                                                                                                                                                                                                                                                                                                                 \\ \hline
\multicolumn{1}{|c|}{ResNet-20}              & \multicolumn{1}{c|}{\multirow{2}{*}{CIFAR-10}}  & \multicolumn{1}{c|}{$98.2\%$} & \multicolumn{1}{c|}{$100\%$} & \multicolumn{1}{c|}{$100\%$} & \multicolumn{1}{c|}{$100\%$} & \multicolumn{1}{c|}{$100\%$} & \multicolumn{1}{c|}{$98.1\%$} & \multicolumn{1}{c|}{$100\%$} & \multicolumn{1}{c|}{$100\%$} & \multicolumn{1}{c|}{$100\%$} & $100\%$ \\ \cline{1-1} \cline{3-12} 
\multicolumn{1}{|c|}{RepVGG-A0}              & \multicolumn{1}{c|}{}                           & \multicolumn{1}{c|}{$99.7\%$} & \multicolumn{1}{c|}{$100\%$} & \multicolumn{1}{c|}{$100\%$} & \multicolumn{1}{c|}{$100\%$} & \multicolumn{1}{c|}{$100\%$} & \multicolumn{1}{c|}{$98.6\%$} & \multicolumn{1}{c|}{$100\%$} & \multicolumn{1}{c|}{$100\%$} & \multicolumn{1}{c|}{$100\%$} & $100\%$ \\ \hline
\multicolumn{1}{|c|}{ResNet-56}              & \multicolumn{1}{c|}{\multirow{2}{*}{CIFAR-100}} & \multicolumn{1}{c|}{$99.7\%$} & \multicolumn{1}{c|}{$100\%$} & \multicolumn{1}{c|}{$100\%$} & \multicolumn{1}{c|}{$100\%$} & \multicolumn{1}{c|}{$100\%$} & \multicolumn{1}{c|}{$97.5\%$} & \multicolumn{1}{c|}{$100\%$} & \multicolumn{1}{c|}{$100\%$} & \multicolumn{1}{c|}{$100\%$} & $100\%$ \\ \cline{1-1} \cline{3-12} 
\multicolumn{1}{|c|}{MobileNet-V2}           & \multicolumn{1}{c|}{}                           & \multicolumn{1}{c|}{$99.7\%$} & \multicolumn{1}{c|}{$100\%$} & \multicolumn{1}{c|}{$100\%$} & \multicolumn{1}{c|}{$100\%$} & \multicolumn{1}{c|}{$100\%$} & \multicolumn{1}{c|}{$99.6\%$} & \multicolumn{1}{c|}{$100\%$} & \multicolumn{1}{c|}{$100\%$} & \multicolumn{1}{c|}{$100\%$} & $100\%$ \\ \hline
\multicolumn{1}{|c|}{InceptionV3}            & \multicolumn{1}{c|}{\multirow{2}{*}{ImageNet-1k}}  & \multicolumn{1}{c|}{$98.2\%$} & \multicolumn{1}{c|}{$100\%$} & \multicolumn{1}{c|}{$100\%$} & \multicolumn{1}{c|}{$100\%$} & \multicolumn{1}{c|}{$100\%$} & \multicolumn{1}{c|}{$100\%$} & \multicolumn{1}{c|}{$100\%$} & \multicolumn{1}{c|}{$100\%$} & \multicolumn{1}{c|}{$100\%$} & $100\%$ \\ \cline{1-1} \cline{3-12} 
\multicolumn{1}{|c|}{DenseNet-201}           & \multicolumn{1}{c|}{}                           & \multicolumn{1}{c|}{$98.7\%$} & \multicolumn{1}{c|}{$100\%$} & \multicolumn{1}{c|}{$100\%$} & \multicolumn{1}{c|}{$100\%$} & \multicolumn{1}{c|}{$100\%$} & \multicolumn{1}{c|}{$99.4\%$} & \multicolumn{1}{c|}{$100\%$} & \multicolumn{1}{c|}{$100\%$} & \multicolumn{1}{c|}{$100\%$} & $100\%$ \\ \hline
\multicolumn{12}{|c|}{Semantic Segmentation (Supervised Learning)}                                                                                                                                                                                                                                                                                                          \\ \hline
\multicolumn{1}{|c|}{U-net}                  & \multicolumn{1}{c|}{Brain-MRI}                  & \multicolumn{1}{c|}{$98.9\%$} & \multicolumn{1}{c|}{$100\%$} & \multicolumn{1}{c|}{$100\%$} & \multicolumn{1}{c|}{$100\%$} & \multicolumn{1}{c|}{$100\%$} & \multicolumn{1}{c|}{$99.2\%$} & \multicolumn{1}{c|}{$100\%$} & \multicolumn{1}{c|}{$100\%$} & \multicolumn{1}{c|}{$100\%$} & $100\%$ \\ \hline
\multicolumn{1}{|c|}{FCN (ResNet-101)}       & \multicolumn{1}{c|}{COCO}                       & \multicolumn{1}{c|}{$98.8\%$} & \multicolumn{1}{c|}{$100\%$} & \multicolumn{1}{c|}{$100\%$} & \multicolumn{1}{c|}{$100\%$} & \multicolumn{1}{c|}{$100\%$} & \multicolumn{1}{c|}{$99.7\%$} & \multicolumn{1}{c|}{$100\%$} & \multicolumn{1}{c|}{$100\%$} & \multicolumn{1}{c|}{$100\%$} & $100\%$ \\ \hline
\multicolumn{12}{|c|}{Generative Method (Unsupervised Learning)}                                                                                                                                                                                                                                                                                                            \\ \hline
\multicolumn{1}{|c|}{DCGAN-D}                  & \multicolumn{1}{c|}{FASHIONGEN}                 & \multicolumn{1}{c|}{$100\%$} & \multicolumn{1}{c|}{$100\%$} & \multicolumn{1}{c|}{$100\%$} & \multicolumn{1}{c|}{$100\%$} & \multicolumn{1}{c|}{$100\%$} & \multicolumn{1}{c|}{$100\%$} & \multicolumn{1}{c|}{$100\%$} & \multicolumn{1}{c|}{$100\%$} & \multicolumn{1}{c|}{$100\%$} & $100\%$ \\ \hline
\end{tabular}
}
\end{table*}

\subsubsection{Fault and Variation-injection Framework}
Several studies have proposed a mathematical model of the non-idealities of the memristor. We use the variation model used in work~\cite{ahmed2022compact, tsai2020robust} that considers spatial and temporal variations and injects random additive and multiplicative Gaussian noise into the weights of pre-trained NNs. To control the severity of the variation, a noise scale $\eta_0$ is used. Similarly, we inject $\mathcal{P}_{flip}$\% of bit- and level-flip-type faults into the weights of pre-trained NNs.

Uncertainty estimation coverage is calculated as:

\begin{equation}\label{eq:coverage}
    \text{coverage} = \frac{\text{\# of } \sigma_y \geq t}{\mathcal{M}}\times 100. 
\end{equation}

Essentially, it calculates the ratio between the number of times the uncertainty of the model $\sigma_y$ is less than a predefined threshold $t$ and the total fault runs $\mathcal{M}$. Also, each injected variation and fault is assumed to impact inference accuracy. We specifically choose noise scales $\eta_0^1\cdots\eta_0^5$ and fault rates $\mathcal{P}_{flip}^1\cdots\mathcal{P}_{flip}^5$ for each model that leads to a degradation of inference accuracy. Specifically, the values of $\eta_0$ and $\mathcal{P}_{flip}$ are chosen between $0.01-0.4$ and $0.02-1.8\%$, respectively. However, subtle accuracy degradation is targeted because the uncertainty in these scenarios is much harder to detect. A Monte Carlo simulation is carried out for each scenario with $\mathcal{M}=1000$ fault runs. For the $t$ value, the uncertainty estimation value $\sigma_y$ of the ideal NN is offset by a small constant $0-0.3$. Therefore, it is stored on hardware alongside the test vector. 




\subsection{Estimating Uncertainty of Variations}\label{sec:var_uncer}

Table~\ref{tab:var} demonstrates the comprehensive evaluation of the coverage of our uncertainty estimation method across a spectrum of NN models under both multiplicative and additive variations with varying noise strengths, denoted by $\eta_0$. Remarkably, our method can consistently achieve $100$\% coverage under most of these diverse conditions. Furthermore, the high coverage percentage across diverse learning paradigms, specifically classification, semantic segmentation, and generative methods, emphasizes the applicability of our method in post-manufacturing and online operations of CIM. Therefore, the reliability under various noise conditions can be maintained to improve the confidence in the prediction.

The input (latent space) of GANs is Gaussian noise. It is equivalent to training an NN with Gaussian noise. As a consequence, the Generator of DCGAN is robust to both types of variation and has low uncertainty. Estimating the uncertainty of a variation robust NN model is unnecessary. Thus, we have not evaluated the uncertainty for this model.

\subsection{Estimating Uncertainty of Bit- and Level-flip}\label{sec:flip_uncer}
Similar to uncertainty due to variations, our method can estimate uncertainty due to bit- and level-flip with consistently $100$\% coverage, as shown in Table~\ref{tab:flip_res}. 
\begin{table*}[]
\caption{
The evaluation of the proposed method in terms of coverage for estimating uncertainty due to both bit- and level-flip faults.
}
\resizebox{\linewidth}{!}{
\begin{tabular}{|cccccccccccc|}
\hline
\multicolumn{1}{|c|}{\multirow{2}{*}{Model}} & \multicolumn{1}{c|}{\multirow{2}{*}{Dataset}}     & \multicolumn{5}{c|}{Bit-flip Faults}                                                                                                                                                                                      & \multicolumn{5}{c|}{Level-flip Faults}                                                                                                                                                               \\ \cline{3-12} 
\multicolumn{1}{|c|}{}                       & \multicolumn{1}{c|}{}                             & \multicolumn{1}{c|}{$\mathcal{P}_{flip}^1$} & \multicolumn{1}{c|}{$\mathcal{P}_{flip^2}$} & \multicolumn{1}{c|}{$\mathcal{P}_{flip}^3$} & \multicolumn{1}{c|}{$\mathcal{P}_{flip}^4$} & \multicolumn{1}{c|}{$\mathcal{P}_{flip}^5$} & \multicolumn{1}{c|}{$\mathcal{P}_{flip}^1$} & \multicolumn{1}{c|}{$\mathcal{P}_{flip}^2$} & \multicolumn{1}{c|}{$\mathcal{P}_{flip}^3$} & \multicolumn{1}{c|}{$\mathcal{P}_{flip}^4$} & $\mathcal{P}_{flip}^5$ \\ \hline \hline
\multicolumn{12}{|c|}{Classification (Supervised Learning)}                                                                                                                                                                                                                                                  \\ \hline
\multicolumn{1}{|c|}{ResNet-20}              & \multicolumn{1}{c|}{\multirow{2}{*}{CIFAR-10}}    & \multicolumn{1}{c|}{$97.2\%$}                & \multicolumn{1}{c|}{$99.0\%$}                & \multicolumn{1}{c|}{$99.7\%$}                & \multicolumn{1}{c|}{$100\%$}                & \multicolumn{1}{c|}{$100\%$}                & \multicolumn{1}{c|}{$98.7\%$}                & \multicolumn{1}{c|}{$99.6\%$}                & \multicolumn{1}{c|}{$99.8\%$}                & \multicolumn{1}{c|}{$99.9\%$}                & $100\%$                \\ \cline{1-1} \cline{3-12}
\multicolumn{1}{|c|}{RepVGG-A0}              & \multicolumn{1}{c|}{}                             & \multicolumn{1}{c|}{$98.9\%$}                & \multicolumn{1}{c|}{$99.6\%$}                & \multicolumn{1}{c|}{$100\%$}                & \multicolumn{1}{c|}{$100\%$}                & \multicolumn{1}{c|}{$99.9\%$}                & \multicolumn{1}{c|}{$99.4\%$}                & \multicolumn{1}{c|}{$99.8\%$}                & \multicolumn{1}{c|}{$100\%$}                & \multicolumn{1}{c|}{$100\%$}                & $100\%$                \\ \hline
\multicolumn{1}{|c|}{ResNet-56}              & \multicolumn{1}{c|}{\multirow{2}{*}{CIFAR-100}}   & \multicolumn{1}{c|}{$96.8\%$}                & \multicolumn{1}{c|}{$99.4\%$}                & \multicolumn{1}{c|}{$100\%$}                & \multicolumn{1}{c|}{$100\%$}                & \multicolumn{1}{c|}{$100\%$}                & \multicolumn{1}{c|}{$98.7\%$}                & \multicolumn{1}{c|}{$99.8\%$}                & \multicolumn{1}{c|}{$100\%$}                & \multicolumn{1}{c|}{$100\%$}                & $100\%$                \\ \cline{1-1} \cline{3-12}
\multicolumn{1}{|c|}{MobileNet-V2}           & \multicolumn{1}{c|}{}                             & \multicolumn{1}{c|}{$99.2\%$}                & \multicolumn{1}{c|}{$99.8\%$}                & \multicolumn{1}{c|}{$100\%$}                & \multicolumn{1}{c|}{$100\%$}                & \multicolumn{1}{c|}{$100\%$}                & \multicolumn{1}{c|}{$99.9\%$}                & \multicolumn{1}{c|}{$100\%$}                & \multicolumn{1}{c|}{$100\%$}                & \multicolumn{1}{c|}{$100\%$}                & $100\%$                \\ \hline
\multicolumn{1}{|c|}{InceptionV3}            & \multicolumn{1}{c|}{\multirow{2}{*}{ImageNet-1k}} & \multicolumn{1}{c|}{$99.6\%$}                & \multicolumn{1}{c|}{$99.9\%$}                & \multicolumn{1}{c|}{$100\%$}                & \multicolumn{1}{c|}{$100\%$}                & \multicolumn{1}{c|}{$100\%$}                & \multicolumn{1}{c|}{$99.9\%$}                & \multicolumn{1}{c|}{$99.9\%$}                & \multicolumn{1}{c|}{$100\%$}                & \multicolumn{1}{c|}{$100\%$}                & $100\%$                \\ \cline{1-1} \cline{3-12} 
\multicolumn{1}{|c|}{DenseNet-201}           & \multicolumn{1}{c|}{}                             & \multicolumn{1}{c|}{$99.5\%$}                & \multicolumn{1}{c|}{$99.9\%$}                & \multicolumn{1}{c|}{$100\%$}                & \multicolumn{1}{c|}{$100\%$}                & \multicolumn{1}{c|}{$100\%$}                & \multicolumn{1}{c|}{$99.8\%$}                & \multicolumn{1}{c|}{$100\%$}                & \multicolumn{1}{c|}{$100\%$}                & \multicolumn{1}{c|}{$100\%$}                & $100\%$                \\ \hline
\multicolumn{12}{|c|}{Semantic Segmentation (Supervised Learning)}                                                                                                                                                                                                                                                                                                                                                                                                                                                                  \\ \hline
\multicolumn{1}{|c|}{U-net}                  & \multicolumn{1}{c|}{Brain-MRI}                    & \multicolumn{1}{c|}{$98.3\%$}                & \multicolumn{1}{c|}{$99.8\%$}                & \multicolumn{1}{c|}{$100\%$}                & \multicolumn{1}{c|}{$100\%$}                & \multicolumn{1}{c|}{$100\%$}                & \multicolumn{1}{c|}{$99.9\%$}                & \multicolumn{1}{c|}{$99.9\%$}                & \multicolumn{1}{c|}{$100\%$}                & \multicolumn{1}{c|}{$100\%$}                & $100\%$                \\ \hline
\multicolumn{1}{|c|}{FCN (ResNet-101)}       & \multicolumn{1}{c|}{COCO}                         & \multicolumn{1}{c|}{$99.0\%$}                & \multicolumn{1}{c|}{$99.3\%$}                & \multicolumn{1}{c|}{$100\%$}                & \multicolumn{1}{c|}{$100\%$}                & \multicolumn{1}{c|}{$100\%$}                & \multicolumn{1}{c|}{$100\%$}                & \multicolumn{1}{c|}{$99.4\%$}                & \multicolumn{1}{c|}{$99.7\%$}                & \multicolumn{1}{c|}{$100\%$}                & $100\%$                \\ \hline
\multicolumn{12}{|c|}{Generative Method (Unsupervised Learning)}                                                                                                                                                                                                                                                                                                                                                                                                                                                                    \\ \hline
\multicolumn{1}{|c|}{DCGAN-Generator}        & \multicolumn{1}{c|}{\multirow{2}{*}{FASHIONGEN}}  & \multicolumn{1}{c|}{$100\%$}                & \multicolumn{1}{c|}{$100\%$}                & \multicolumn{1}{c|}{$100\%$}                & \multicolumn{1}{c|}{$100\%$}                & \multicolumn{1}{c|}{$100\%$}                & \multicolumn{1}{c|}{$100\%$}                & \multicolumn{1}{c|}{$100\%$}                & \multicolumn{1}{c|}{$100\%$}                & \multicolumn{1}{c|}{$100\%$}                & $100\%$                \\ \cline{1-1} \cline{3-12} 
\multicolumn{1}{|c|}{DCGAN-Discriminator}    & \multicolumn{1}{c|}{}                             & \multicolumn{1}{c|}{$100\%$}                & \multicolumn{1}{c|}{$100\%$}                & \multicolumn{1}{c|}{$100\%$}                & \multicolumn{1}{c|}{$100\%$}                & \multicolumn{1}{c|}{$100\%$}                & \multicolumn{1}{c|}{$100\%$}                & \multicolumn{1}{c|}{$100\%$}                & \multicolumn{1}{c|}{$100\%$}                & \multicolumn{1}{c|}{$100\%$}                & $100\%$                \\ \hline
\end{tabular}
}
\label{tab:flip_res}
\end{table*}
Even though, the Generator of DCGAN is robust to variations, it is susceptible to bit- and level-flip-type faults. Therefore, we have evaluated its uncertainty.

\subsection{Verifiable Uncertainty Estimates}
To increase the confidence in the uncertainty estimates, we have modified equation~\ref{eq:coverage} into:

\begin{equation}\label{eq:coverage_varifyable}
    \text{coverage} = \frac{\text{\# of } (\sigma_y \geq t \textit{ and } A_{test} < A_{baseline})}{\mathcal{M}}\times 100. 
\end{equation}
\begin{table*}[]
\caption{Evaluation of the uncertainty estimation coverage with accuracy degradation verified in each step. The same noise scales $\eta_0$ and fault rates $\mathcal{P}_{flip}$ are used as in Tables~\ref{tab:var} and~\ref{tab:flip_res}. }
\resizebox{\linewidth}{!}{
\begin{tabular}{|c|c|ccccc|ccccc|}
\hline
\multirow{2}{*}{Model} & \multirow{2}{*}{Dataset}   & \multicolumn{5}{c|}{Multiplicative Variations}                                                                                                                                                                 & \multicolumn{5}{c|}{Additive Variations}                                                                                                                                                                       \\ \cline{3-12} 
                       &                            & \multicolumn{1}{c|}{$\eta_0^1$}             & \multicolumn{1}{c|}{$\eta_0^2$}             & \multicolumn{1}{c|}{$\eta_0^3$}             & \multicolumn{1}{c|}{$\eta_0^4$}             & $\eta_0^5$             & \multicolumn{1}{c|}{$\eta_0^1$}             & \multicolumn{1}{c|}{$\eta_0^2$}             & \multicolumn{1}{c|}{$\eta_0^3$}             & \multicolumn{1}{c|}{$\eta_0^4$}             & $\eta_0^5$             \\ \hline
ResNet-20              & \multirow{2}{*}{CIFAR-10}  & \multicolumn{1}{c|}{$100\%$}                & \multicolumn{1}{c|}{$100\%$}                & \multicolumn{1}{c|}{$100\%$}                & \multicolumn{1}{c|}{$100\%$}                & $100\%$                & \multicolumn{1}{c|}{$100\%$}                & \multicolumn{1}{c|}{$100\%$}                & \multicolumn{1}{c|}{$100\%$}                & \multicolumn{1}{c|}{$100\%$}                & $100\%$                \\ \cline{1-1} \cline{3-12} 
RepVGG-A0              &                            & \multicolumn{1}{c|}{$100\%$}                & \multicolumn{1}{c|}{$100\%$}                & \multicolumn{1}{c|}{$100\%$}                & \multicolumn{1}{c|}{$100\%$}                & $100\%$                & \multicolumn{1}{c|}{$100\%$}                & \multicolumn{1}{c|}{$100\%$}                & \multicolumn{1}{c|}{$100\%$}                & \multicolumn{1}{c|}{$100\%$}                & $100\%$                \\ \hline
ResNet-56              & \multirow{2}{*}{CIFAR-100} & \multicolumn{1}{c|}{$99.1\%$}                & \multicolumn{1}{c|}{$100\%$}                & \multicolumn{1}{c|}{$100\%$}                & \multicolumn{1}{c|}{$100\%$}                & $100\%$                & \multicolumn{1}{c|}{$99.9\%$}                & \multicolumn{1}{c|}{$100\%$}                & \multicolumn{1}{c|}{$100\%$}                & \multicolumn{1}{c|}{$100\%$}                & $100\%$                \\ \cline{1-1} \cline{3-12} 
MobileNet-V2           &                            & \multicolumn{1}{c|}{$99.7\%$}                & \multicolumn{1}{c|}{$100\%$}                & \multicolumn{1}{c|}{$100\%$}                & \multicolumn{1}{c|}{$100\%$}                & $100\%$                & \multicolumn{1}{c|}{$99.1\%$}                & \multicolumn{1}{c|}{$100\%$}                & \multicolumn{1}{c|}{$100\%$}                & \multicolumn{1}{c|}{$100\%$}                & $100\%$                \\ \hline
\multirow{2}{*}{Model} & \multirow{2}{*}{Dataset}   & \multicolumn{5}{c|}{Bit-flip Faults}                                                                                                                                                                           & \multicolumn{5}{c|}{Level-flip Faults}                                                                                                                                                                         \\ \cline{3-12} 
                       &                            & \multicolumn{1}{c|}{$\mathcal{P}_{flip}^1$} & \multicolumn{1}{c|}{$\mathcal{P}_{flip}^2$} & \multicolumn{1}{c|}{$\mathcal{P}_{flip}^3$} & \multicolumn{1}{c|}{$\mathcal{P}_{flip}^4$} & $\mathcal{P}_{flip}^5$ & \multicolumn{1}{c|}{$\mathcal{P}_{flip}^1$} & \multicolumn{1}{c|}{$\mathcal{P}_{flip}^2$} & \multicolumn{1}{c|}{$\mathcal{P}_{flip}^3$} & \multicolumn{1}{c|}{$\mathcal{P}_{flip}^4$} & $\mathcal{P}_{flip}^5$ \\ \hline
ResNet-20              & \multirow{2}{*}{CIFAR-10}  & \multicolumn{1}{c|}{$100\%$}                & \multicolumn{1}{c|}{$100\%$}                & \multicolumn{1}{c|}{$100\%$}                & \multicolumn{1}{c|}{$100\%$}                & $100\%$                & \multicolumn{1}{c|}{$100\%$}                & \multicolumn{1}{c|}{$100\%$}                & \multicolumn{1}{c|}{$100\%$}                & \multicolumn{1}{c|}{$100\%$}                & $100\%$                \\ \cline{1-1} \cline{3-12} 
RepVGG-A0              &                            & \multicolumn{1}{c|}{$100\%$}                & \multicolumn{1}{c|}{$100\%$}                & \multicolumn{1}{c|}{$100\%$}                & \multicolumn{1}{c|}{$100\%$}                & $100\%$                & \multicolumn{1}{c|}{$100\%$}                & \multicolumn{1}{c|}{$100\%$}                & \multicolumn{1}{c|}{$100\%$}                & \multicolumn{1}{c|}{$100\%$}                & $100\%$                \\ \hline
ResNet-56              & \multirow{2}{*}{CIFAR-100} & \multicolumn{1}{c|}{$96.8\%$}                & \multicolumn{1}{c|}{$99.4\%$}                & \multicolumn{1}{c|}{$100\%$}                & \multicolumn{1}{c|}{$100\%$}                & $100\%$                & \multicolumn{1}{c|}{$99.4\%$}                & \multicolumn{1}{c|}{$100\%$}                & \multicolumn{1}{c|}{$100\%$}                & \multicolumn{1}{c|}{$100\%$}                & $100\%$                \\ \cline{1-1} \cline{3-12} 
MobileNet-V2           &                            & \multicolumn{1}{c|}{$98.2\%$}                & \multicolumn{1}{c|}{$99.8\%$}                & \multicolumn{1}{c|}{$100\%$}                & \multicolumn{1}{c|}{$100\%$}                & $100\%$                & \multicolumn{1}{c|}{$99.7\%$}                & \multicolumn{1}{c|}{$100\%$}                & \multicolumn{1}{c|}{$100\%$}                & \multicolumn{1}{c|}{$100\%$}                & $100\%$                \\ \hline
\end{tabular}
}
\label{tab:verifiable}
\end{table*}

This allows us to calculate \emph{verifiable coverage} for uncertainty estimates. Here, $A_{test}$ represents the inference accuracy after a fault or variation injection, and $A_{baseline}$ is the baseline inference accuracy. However, due to computational limitations, we have only evaluated this approach on a subset of the overall models shown in Table~\ref{tab:models}.

As depicted in Table~\ref{tab:verifiable}, our uncertainty estimation approach can still achieve $100\%$ coverage (most of the time) for uncertainty estimates in different fault rates and variations. This further underscores the robustness of our approach.

\subsection{Layer-wise Uncertainty Estimation}
We have extensively evaluated our approach when all the parameters of NN are affected by the memristive non-idealities. However, it is likely that not all the layers in an NN are affected by those non-idealities. Therefore, we have done further evaluations of our approach by randomly injecting faults and variations into $10-50\%$ of the layers of the CIFAR-10 and CIFAR-100 models. As demonstrated in Table~\ref{tab:layer_wise}, our method can achieve $100\%$ uncertainty estimation coverage when $\geq 20\%$ of the layers of NNs are affected by variations or faults. Even when only $10\%$ of the layers are affected, the uncertainty due to faults or variations that lead to $\geq 1-2\%$ accuracy degradation can be estimated with our approach.

\subsection{Analysis of the Impact of Threshold Value on Coverage}
Choosing the right value for the threshold $t$ is important to achieve high uncertainty estimation coverage. It implicitly reduces the risk of false positive or negative uncertainty estimates. We perform a series of analyses with different offsets for $t$. As demonstrated in Table~\ref{tab:offset_analysis}, as the offset increases, the coverage gradually decreases to a close value $0\%$. Therefore, it is beneficial to use the threshold $t$ the same as the SD of the ideal MUT. However, $t$ should never be chosen below the SD of the ideal MUT. In this case, coverage could be high due to false-positive uncertainty estimation.
\begin{table*}[]
\caption{Evaluation of the coverage of the proposed uncertainty estimate approach with faults and variations injected into a random subset (10-50\%) of all layers. Here, the fault rate and the noise scale $\eta_0$ are kept constant. }
\resizebox{\linewidth}{!}{
\begin{tabular}{|c|c|ccccc|ccccc|}
\hline
\multirow{2}{*}{Model} & \multirow{2}{*}{Dataset}   & \multicolumn{5}{c|}{Multiplicative Variations}                                                                                      & \multicolumn{5}{c|}{Additive Variations}                                                                                            \\ \cline{3-12} 
                       &                            & \multicolumn{1}{c|}{$10\%$}  & \multicolumn{1}{c|}{$20\%$}  & \multicolumn{1}{c|}{$30\%$}  & \multicolumn{1}{c|}{$40\%$}  & $50\%$  & \multicolumn{1}{c|}{$10\%$}  & \multicolumn{1}{c|}{$20\%$}  & \multicolumn{1}{c|}{$30\%$}  & \multicolumn{1}{c|}{$40\%$}  & $50\%$  \\ \hline\hline
ResNet-20              & \multirow{2}{*}{CIFAR-10}  & \multicolumn{1}{c|}{$97.8\%$} & \multicolumn{1}{c|}{$100\%$} & \multicolumn{1}{c|}{$100\%$} & \multicolumn{1}{c|}{$100\%$} & $100\%$ & \multicolumn{1}{c|}{$88.8\%$} & \multicolumn{1}{c|}{$99.7\%$} & \multicolumn{1}{c|}{$100\%$} & \multicolumn{1}{c|}{$100\%$} & $100\%$ \\ \cline{1-1} \cline{3-12} 
RepVGG-A0              &                            & \multicolumn{1}{c|}{$100\%$} & \multicolumn{1}{c|}{$100\%$} & \multicolumn{1}{c|}{$100\%$} & \multicolumn{1}{c|}{$100\%$} & $100\%$ & \multicolumn{1}{c|}{$100\%$} & \multicolumn{1}{c|}{$100\%$} & \multicolumn{1}{c|}{$100\%$} & \multicolumn{1}{c|}{$100\%$} & $100\%$ \\ \hline
ResNet-56              & \multirow{2}{*}{CIFAR-100} & \multicolumn{1}{c|}{$96.9\%$} & \multicolumn{1}{c|}{$100\%$} & \multicolumn{1}{c|}{$100\%$} & \multicolumn{1}{c|}{$100\%$} & $100\%$ & \multicolumn{1}{c|}{$96.3\%$} & \multicolumn{1}{c|}{$99.8\%$} & \multicolumn{1}{c|}{$100\%$} & \multicolumn{1}{c|}{$100\%$} & $100\%$ \\ \cline{1-1} \cline{3-12} 
MobileNet-V2           &                            & \multicolumn{1}{c|}{$100\%$} & \multicolumn{1}{c|}{$100\%$} & \multicolumn{1}{c|}{$100\%$} & \multicolumn{1}{c|}{$100\%$} & $100\%$ & \multicolumn{1}{c|}{$99.7\%$} & \multicolumn{1}{c|}{$100\%$} & \multicolumn{1}{c|}{$100\%$} & \multicolumn{1}{c|}{$100\%$} & $100\%$ \\ \hline
                       &                            & \multicolumn{5}{c|}{Bit-flip Faults}                                                                                                & \multicolumn{5}{c|}{Level-flip Faults}                                                                                              \\ \hline
ResNet-20              & \multirow{2}{*}{CIFAR-10}  & \multicolumn{1}{c|}{$95.2\%$} & \multicolumn{1}{c|}{$99.7\%$} & \multicolumn{1}{c|}{$100\%$} & \multicolumn{1}{c|}{$100\%$} & $100\%$ & \multicolumn{1}{c|}{$96.2\%$} & \multicolumn{1}{c|}{$97.0\%$} & \multicolumn{1}{c|}{$96.7\%$} & \multicolumn{1}{c|}{$96.4\%$} & $97.6\%$ \\ \cline{1-1} \cline{3-12} 
RepVGG-A0              &                            & \multicolumn{1}{c|}{$97.3\%$} & \multicolumn{1}{c|}{$100\%$} & \multicolumn{1}{c|}{$100\%$} & \multicolumn{1}{c|}{$100\%$} & $100\%$ & \multicolumn{1}{c|}{$100\%$} & \multicolumn{1}{c|}{$100\%$} & \multicolumn{1}{c|}{$100\%$} & \multicolumn{1}{c|}{$100\%$} & $100\%$ \\ \hline
ResNet-56              & \multirow{2}{*}{CIFAR-100} & \multicolumn{1}{c|}{$98.0\%$} & \multicolumn{1}{c|}{$99.6\%$} & \multicolumn{1}{c|}{$100\%$} & \multicolumn{1}{c|}{$100\%$} & $100\%$ & \multicolumn{1}{c|}{$100\%$} & \multicolumn{1}{c|}{$100\%$} & \multicolumn{1}{c|}{$100\%$} & \multicolumn{1}{c|}{$100\%$} & $100\%$ \\ \cline{1-1} \cline{3-12} 
MobileNet-V2           &                            & \multicolumn{1}{c|}{$99.8\%$} & \multicolumn{1}{c|}{$100\%$} & \multicolumn{1}{c|}{$100\%$} & \multicolumn{1}{c|}{$100\%$} & $100\%$ & \multicolumn{1}{c|}{$100\%$} & \multicolumn{1}{c|}{$100\%$} & \multicolumn{1}{c|}{$100\%$} & \multicolumn{1}{c|}{$100\%$} & $100\%$ \\ \hline
\end{tabular}
}
\label{tab:layer_wise}
\end{table*}

\subsection{Resolution of Uncertainty Estimation}
As mentioned previously, we have used a minuscule noise scale $\eta_0$ and $P_{flip}$, leading to negligible inference accuracy loss. Although our method can achieve $100\%$ coverage for uncertainty estimates in those scenarios, it is important to find the boundary of coverage to determine the risk of false positive uncertainty estimates. Therefore, we have conducted several experiments on the CIFAR-10 and CIFAR-100 datasets, with even lower $P_{flip}$ and noise scales. We have found that our method can estimate uncertainty with $100\%$ coverage, even with $1-2\%$ accuracy degradation. However,
when the accuracy degradation is very low, e.g., $\leq 0.5\%$, the number of false-negative uncertainty estimates can be as high as $5\%$.

\begin{table*}[]
\caption{The effect of offset value of $t$ on uncertainty estimation coverage. Evaluated on multiplicative variations with the same noise scale $\eta_0$ as Table~\ref{tab:var}.}
\centering
\footnotesize
\begin{tabular}{|c|c|c|c|c|c|}
\hline
\multirow{2}{*}{\begin{tabular}[c]{@{}c@{}}Offset of \\ Threshold $t$\end{tabular}} & Model    & \multirow{2}{*}{\begin{tabular}[c]{@{}c@{}}Offset of \\ Threshold $t$\end{tabular}} & Model     & \multirow{2}{*}{\begin{tabular}[c]{@{}c@{}}Offset of \\ Threshold $t$\end{tabular}} & Model        \\ \cline{2-2} \cline{4-4} \cline{6-6} 
                                                                                    & RepVGG   &                                                                                     & ResNet-56 &                                                                                     & DenseNet-201 \\ \hline
0.0                                                                                 & $100\%$  & 0.0                                                                                 & $100\%$   & 0.000                                                                               & $100\%$      \\ \hline
0.40                                                                                & $100\%$  & 0.140                                                                               & $100\%$   & 0.040                                                                               & $100\%$      \\ \hline
0.42                                                                                & $99.9\%$ & 0.150                                                                               & $99.7\%$  & 0.050                                                                               & $93.5\%$     \\ \hline
0.43                                                                                & $97.0\%$ & 0.155                                                                               & $94.4\%$  & 0.051                                                                               & $77.9\%$     \\ \hline
0.44                                                                                & $66.2\%$ & 0.160                                                                               & $74.0\%$  & 0.052                                                                               & $50.3\%$     \\ \hline
0.45                                                                                & $18.6\%$ & 0.165                                                                               & $42.8\%$  & 0.053                                                                               & $22.2\%$     \\ \hline
0.46                                                                                & $1.8\%$  & $0.170$                                                                               & $17.1\%$  & 0.054                                                                               & $5.8\%$      \\ \hline
\end{tabular}
\label{tab:offset_analysis}
\end{table*}

\subsection{Comparison With Related Works}\label{sec:comp_rel}
We compare our approach to the related works with point estimate test vectors and Bayesian optimized test vectors, which employ a functional approach. Our single Bayesian test vector outperforms the methods proposed by Chen et al.~\cite{chen2021line}, Li et al.~\cite{li2019rramedy}, Luo et al.~\cite{luo2019functional}, and Ahmed et al.~\cite{ahmed2022compact} on all metrics, as displayed in Table~\ref{tab:comp_rel}, even though we use only a single test vector, test query and forward pass. This implies that our method significantly reduces latency and energy consumption for uncertainty estimation compared to the other methods. The latency and the energy consumption are directly proportional to the number of test vectors used for estimating uncertainty. For instance, the testing method of Li et al.~\cite{li2019rramedy} requires $1024$ test vectors, thereby necessitating $1024$ times more matrix-vector multiplication operations and power consumption. In our comparisons, we assume that all methods employ identical hardware implementation, NN topology, and NVM technology.

\begin{table*}
    \centering
    \footnotesize
    \caption{Comparison of the proposed approach with the existing methods using different performance metrics. To ensure a fair comparison, the analysis of all approaches is conducted on the CIFAR-10 dataset. The memory consumptions for the test vectors and their labels are calculated based on the bit width reported by~\cite{chen2021line}.
    }
    \begin{threeparttable}
\begin{tabular}{|c|c|c|c|c|c|c|}
    \hline 
    \multicolumn{1}{|c|}{Methods} & \cite{chen2021line} & \cite{li2019rramedy} & \cite{luo2019functional} & \cite{ahmed2022compact} & \cite{Arjun2023BO} & \multicolumn{1}{c|}{Proposed}  \tabularnewline
    \hline 
    \multicolumn{1}{|c|}{Size of } & \multirow{2}{*}{$10000$} & \multirow{2}{*}{$1024$} & \multirow{2}{*}{${10}$-${50}$} & \multirow{2}{*}{$16$-$64$} & \multirow{2}{*}{$9$\tnote{v}~/$1$\tnote{vi}} & \multirow{2}{*}{$\bm{1}$} \tabularnewline
    \multicolumn{1}{|c|}{test vectors} &  &  &  & & & \multicolumn{1}{c|}{} \tabularnewline
    \hline 
    \# of test queries  & \multirow{2}{*}{$10000$} & \multirow{2}{*}{$1024$} & \multirow{2}{*}{$10$-$50$} & \multirow{2}{*}{${17}$} & \multirow{2}{*}{$9$\tnote{v}~/$1$\tnote{vi}}  & \multirow{2}{*}{$\bm{1}$} \tabularnewline
    (normalized)&  &  &  & & & \multicolumn{1}{c|}{} \tabularnewline
    \hline 
    Memory & $0.015$\tnote{i} & $234.42$\tnote{i} & $0.154$\tnote{i} & ${0.0003}$\tnote{i} & \begin{tabular}{@{}c@{}}$0.1105$\tnote{v} \\ /$0.0123$\tnote{vi}\end{tabular} & \multicolumn{1}{c|}{$\bm{0.02455}$\tnote{i}} \tabularnewline
    \cline{2-7}
    overhead (MB) & $245.775$\tnote{ii} & $234.42$\tnote{ii} & $0.154$\tnote{ii} & ${0.1966}$\tnote{ii} & \begin{tabular}{@{}c@{}}$0.1105$\tnote{v} \\ /$0.0123$\tnote{vi}\end{tabular} & \multicolumn{1}{c|}{$\bm{0.02455}$\tnote{ii}} \tabularnewline
    \hline 
    \multirow{1}{*}{Coverage (\%)} & $99.27$ & $98$ & $76$ / $84$ & ${100}$ & $86.7$\tnote{v}~/$30$\tnote{vi}  & \multicolumn{1}{c|}{$\bm{100}$} \tabularnewline
    \hline 
\end{tabular}
    \begin{tablenotes}
    \item[i] Re-training data is stored in hardware.
    \item[ii] Re-training data is not stored in hardware.
    \item[V] 32-bit floating-point, Re-training data is stored in hardware.
    \item[Vi] 16-bit floating-point, Re-training data is not stored in hardware.
    \end{tablenotes}
    \end{threeparttable}
    \label{tab:comp_rel}
\end{table*}

Moreover, our proposed approach consistently achieves $100\%$ coverage in various fault and variation scenarios, exceeding the coverage rates of $30\%$ to $99.27\%$ achieved by the methods proposed by Chen et al.~\cite{chen2021line}, Li et al.~\cite{li2019rramedy}, Luo et al.~\cite{luo2019functional}, and A. Chaudhuri et al.~\cite{Arjun2023BO}. 

In terms of the storage overhead of the Bayesian test vector, our method requires only $0.0245$ MB to store the test vector, which is equivalent to storing only two (point estimate) test vectors. This is because our method requires the storage of mean and variance elementwise. Nevertheless, the memory requirement of our method is substantially lower than the other methods, regardless of whether re-training data is stored in hardware. Importantly, our method does not depend on storing re-training data in hardware to reduce memory consumption, unlike the methods proposed by Chen et al.~\cite{chen2021line} and Ahmed et al.~\cite{ahmed2022compact}.

In terms of sampling overhead for uncertainty estimation, there is a trade-off between the number of MC samples required $N$, the number of neurons in the penultimate layer of MUT $C$, and the storage overhead. For a MUT with a larger $C$, e.g., $C\geq10$, only one sample is required, $N=1$. In this instance, it is beneficial to take one MC sample before NN is deployed to CIM and store the sample on the hardware. Consequently, the number of MC samplings is reduced to one for the entire device operation, and the storage requirement for the Bayesian test vector is reduced to $0.0123$ MB, a reduction of $2\times$. Overall, the overhead is the same as that of one test run for related methods. On the other hand, for an MUT with a small $C$, e.g., $C=1~\textit{or}~2$, sampling overhead can be reduced by taking multiple samples and storing them. In this case, the sampling overhead is reduced by $N\times$ in each of the following uncertainty estimation steps, but the storage overhead increases by $N\times$. For an edge device with limited memory, storing the parameters of a Bayesian test vector and taking samples in each uncertainty estimation step is more beneficial. In addition, the number of elements in an input determines how many E-samples are required in an uncertainty estimation step. 
Nevertheless, Elementwise sampling can be done in parallel since the input, e.g., an image, is a 3-D matrix.



\section{Conclusion}\label{sec:conclusion}
In this work, we propose a single Bayesian test vector generation framework to estimate the uncertainty of memristive NNs. The proposed Bayesian test vector is specifically optimized to provide low uncertainty output for fault- and variation-free memristive NNs. Our method requires only single (element-wise) sampling from the distribution of the Bayesian test vector and a single forward pass on a larger model. Thus, the overhead associated with our approach is minimal. In addition, we proposed an application of our uncertainty estimation approach in the pre-deployment and post-deployment scenarios of an NN to MHA. We have consistently demonstrated high uncertainty estimation coverage of our approach on various NN topologies, tasks, fault rates, and noise scales related to variations. Our work improves confidence in the predictions made by memristive NNs.

\bibliographystyle{IEEEtran}
\typeout{}
\bibliography{references}

\end{document}